%% file: ms.tex
\documentclass{IOS-Book-Article}

\usepackage{mathptmx}
\usepackage[T1]{fontenc}
\usepackage[utf8]{inputenc}
\usepackage{times}
\usepackage{graphicx}
\usepackage{amsmath}
\usepackage{flexisym}
\usepackage{color}
\usepackage{makecell}
\usepackage{booktabs}
\usepackage{subcaption}
\usepackage{microtype}

\def\hb{\hbox to 10.7 cm{}}

\begin{document}

\pagestyle{headings}
\def\thepage{}

\begin{frontmatter}              

\title{Neural Morphological Tagging for Estonian}

\markboth{}{}

\author{\fnms{Alexander} \snm{Tkachenko}
\thanks{Corresponding Author: Alexander Tkachenko; E-mail: aleksandr.tkatsenko@ut.ee.}} and
\author{\fnms{Kairit} \snm{Sirts}}
\runningauthor{A. Tkachenko et al.}
\address{Institute of Computer Science, University of Tartu, Tartu, Estonia}

\begin{abstract}
We develop neural morphological tagging and disambiguation models for Estonian. First, we experiment with two neural architectures for morphological tagging -- a standard multiclass classifier which treats each morphological tag as a single unit, and a sequence model which handles the morphological tags as sequences of morphological category values.
Secondly, we complement these models with the analyses generated by a rule-based Estonian morphological analyser (MA) \textsc{Vabamorf}, thus performing a soft morphological disambiguation.
We compare two ways of supplementing a neural morphological tagger with the MA outputs: firstly, by adding the combined analyses embeddings to the word representation input to the neural tagging model, and secondly, by adopting an attention mechanism to focus on the most relevant analyses generated by the MA.
Experiments on three Estonian datasets show that our neural architectures consistently outperform the non-neural baselines, including HMM-disambiguated \textsc{Vabamorf}, while augmenting models with MA outputs results in a further performance boost for both models.
\end{abstract}

\end{frontmatter}
\markboth{}{}

\input{introduction}

\vspace{-1em}
\input{vabamorf}

\vspace{-1em}
\input{models}
\vspace{-1em}
\input{incorporating_ma}
\input{experiments}
\input{results}

\input{conclusion}
\input{acknowledgments}

\bibliography{paper}
\bibliographystyle{ieeetr}

\end{document}

%% file: introduction.tex
\section{Introduction}
We address the problem of morphological tagging of Estonian language.
Existing resources for Estonian computational morphological processing include a rule-based morphological analyser \textsc{Vabamorf}~\cite{Kaalep1997} which generates all possible morphological analyses for a word. \textsc{Vabamorf} also includes a statistical HMM-based disambiguator which, based on the context, resolves the ambiguities for most of the words \cite{kaalep2001}. 

Recently, neural models have been successfully applied to both POS and morphological tagging achieving prominent results in many languages~\cite{plank2016,yu2017,heigold2017}. In this paper, we focus on neural morphological tagging of Estonian. Among other languages, Heigold et al.~\cite{heigold2017}, and Tkachenko and Sirts~\cite{tkachenko2018} also report morphological tagging results for Estonian trained on Universal Dependencies (UD) corpus, reaching 94.25\% and 93.30\% respectively on different versions of the UD Estonian test set.



We experiment with two neural architectures which differ in the way they model morphological tags.
The first architecture is a standard multiclass classifier which treats each morphological tag as a single unit.
As a second architecture, we adopt the sequential architecture proposed by Tkachenko and Sirts~\cite{tkachenko2018} that models each morphological tag as a sequence of morphological categories.
Both architectures use a bidirectional LSTM encoder to compute the representations of words  and their contexts. 

Morphological disambiguation is a task closely related to morphological tagging. While the goal of morphological tagging is to select the correct label from the set of all possible tags, the task of a morphological disambiguator is to select the correct analysis from a limited set of candidates provided by a morphological analyser (MA)~\cite{Shen2016,Zalmout2017}.

The main drawback of morphological disambiguation is that it totally relies on MA. If the MA output does not contain the correct tag (or in some cases, any tag at all), the model will not be able to make the correct prediction.
Furthermore, the straightforward morphological disambiguation is complicated or inapplicable in situations where the available annotated dataset and the MA use different tag annotation schemes.

To address these issues, we experiment with two ways to utilise MA outputs as additional inputs to our neural models, thus performing a soft morphological disambiguation \cite{inoue2017}.
The first approach, inspired by the one proposed by Inoue et al.~\cite{inoue2017}, combines the embedded representations of the word's morphological analyses and feeds this combined representation, in addition to the representation of the word itself, as input to the neural tagging model. The second architecture again embeds the morphological analyses generated by the analyser but adopts an attention mechanism -- a method commonly used in neural machine translation models \cite{Bahdanau2014,Luong2015} -- to focus on the most relevant analyses when making the tagging decisions.

We evaluate our models on three different Estonian morphologically annotated datasets.
Both neural models demonstrate high performance, reaching up to 97\% of full tag accuracy and consistently outperforming both the rule-based \textsc{Vabamorf} with statistical disambiguation, and a strong CRF-based baseline.
By further augmenting our models with the outputs of the \textsc{Vabamorf} MA, we achieve performance gains in the range of 0.46\% to 1.15\% in the absolute full tag accuracy for all models on all datasets.
Furthermore, we show that incorporating the outputs of the MA is beneficial even if the analyser and the training dataset use different annotation schemes.


%% file: vabamorf.tex
\section{Vabamorf}
\label{sec:vabamorf}

We start by analysing the tagging accuracy of the \textsc{Vabamorf} MA.
\textsc{Vabamorf} is a publicly available morphological analyser for Estonian language which is available on its own\footnote{https://github.com/Filosoft/vabamorf} and is also included in the EstNLTK  toolkit\footnote{https://estnltk.github.io/estnltk} \cite{orasmaa2016}.
It processes text in two steps.
First, it performs rule-based morphological analysis on each individual word in order to identify their possible analyses.
Since very often multiple analyses are produced, as a second step, \textsc{Vabamorf} employs a HMM-based disambiguator to identify words' correct analyses based on the context.



However, \textsc{Vabamorf} is not perfect and introduces errors in both steps. 
According to Kaalep and Vaino \cite{kaalep2001}, for ca 0.4\% of words in the texts, the analyser produces a list of analyses that does not contain the correct analysis.
Furthermore, for about 13.5\% of the cases, the disambiguator purposely avoids resolving ambiguities and outputs multiple analyses.  While such behaviour might be appropriate for a corpus linguistic study of the text, it is unsuitable when morphological disambiguation is adopted as an intermediate step in a natural language processing pipeline.

To assess the magnitude of these errors, we analysed the performance of \textsc{Vabamorf} on one of our experimental datasets -- the Estonian morphologically disambiguated corpus (\textsc{MD}). The statistics of this corpus are given in Table~\ref{tbl:corpus_stats}.
We ran \textsc{Vabamorf} with the guesser and proper name resolver on the whole MD dataset. 
Results show that for 8.4\% of words, \textsc{Vabamorf} retains ambiguity on MD dataset.



Since \textsc{Vabamorf} and the \textsc{MD} dataset use different tagsets, it is impossible to evaluate the disambiguation performance of \textsc{Vabamorf} directly.
Therefore, we convert \textsc{Vabamorf} outputs to the \textsc{MD}-compatible format using a set of scripts\footnote{https://github.com/EstSyntax/EstCG/blob/master/konverter.sh} that perform required transformations and on top of that run a constraint grammar parser.
As a side effect, such conversion introduces additional ambiguity of 4.22\% compared to the \textsc{Vabamorf}'s original output.
We further post-process the obtained output as well as the \textsc{MD} dataset in order to eliminate some systematic inconsistencies in annotations for pronouns, capitalised words, etc.
As a result, 12.62\% of words remain ambiguous, 97.07\% of which have the correct analysis  in the list. 
In the case of unambiguous words, the accuracy is 97.41\%.

In order to assess the overall accuracy, we need to resolve the ambiguous cases.
We implemented two simple ways to do so: by choosing either a random or the first analysis out of the list of candidates.
This resulted in an overall accuracy of 89.76\% and 90.29\% respectively.
We conclude that unresolved ambiguity significantly affects \textsc{Vabamorf}'s overall performance and complicates its direct application to the task of morphological tagging in a NLP pipeline.

%% file: models.tex
\section{Morphological Tagging Models}
\label{sec:tagging}

We experiment with two different neural architectures for morphological tagging: a multiclass model which treats morphological labels as a single tag, and a sequence model which generates morphological labels as sequences of feature-value pairs. 
Both models utilise the general encoder-decoder neural architecture where the encoder computes the representations for the inputs and the decoder performs the labelling based on these input representations.
We first describe the encoder architecture which is the same for all models. Then, we describe the decoder architectures.



\paragraph{Encoder}
\label{sec:encoder}
We adopt a standard neural sequence tagging encoder architecture for both models.
Similar encoder architectures have been applied recently with notable success to morphological tagging \cite{heigold2017,yu2017,tkachenko2018} as well as several other sequence tagging tasks \cite{lample2016,chiu2016,Ling2015}.
It consists of a  bidirectional LSTM network that maps words in a sentence into contextual feature vectors using character and word-level embeddings.
Character-level word embeddings $\mathbf{w}_{\textbf{char}}$ are constructed with another bidirectional LSTM network and they capture useful information about words' morphology and shape.
Word-level embeddings $\mathbf{w}$ are initialised with pre-trained embeddings and fine-tuned during training.
The character and word-level embeddings are concatenated and passed as inputs $\mathbf{\bar{w}}$ to the bidirectional LSTM encoder.
The resulting hidden states $\mathbf{h}$ capture contextual information for each word in a sentence.


\paragraph{Multiclass Model (\textsc{Mc})}
The first decoder employs the standard multiclass classifier used by both Heigold et al.~\cite{heigold2017} and Yu et al.~\cite{yu2017}.
It is essentially just a softmax classifier over the full tagset.
The inherent limitation of this model is the inability to predict tags that are not present in the training corpus. 


\paragraph{Sequence Model (\textsc{Seq})}
The sequence model, introduced by Tkachenko and Sirts \cite{tkachenko2018}, predicts complex morphological labels as sequences of category values.
For each word in a sentence, the sequential decoder generates a sequence of morphological feature-value pairs based on the context vector $\mathbf{h}$ and the previous predictions.
The decoder is a unidirectional LSTM network.
Decoding starts by passing the start-of-sequence symbol as input.
At each time step, the decoder computes the label context vector $\mathbf{g}_j$ based on the embedding of the previously predicted category value $\mathbf{f}_{j-1}$, previous label context vector $\mathbf{g}_{j-1}$ and the word's context vector $\mathbf{h}$.
The probability of each morphological feature-value pair is then computed with a softmax classifier over all possible feature-value pairs.
At training time, we feed correct labels as inputs while at inference time, we greedily emit the best prediction.
The decoding terminates once the end-of-sequence symbol is produced.

%% file: incorporating_ma.tex
\section{Incorporating Morphological Analyser}
\label{sec:disambiguation}


We explore two methods to incorporate the outputs generated by the \textsc{Vabamorf} MA.
In both cases, we encode the MA outputs using dense vector representations. We make use of the following notation. For a word $w$, the MA generates $M$ distinct morphological analyses $m_1, \dots, m_M$. Each of these morphological analyses can be split into a list of  $C$ category-value pairs $m = [f_1, \dots, f_C]$. The embedding vectors corresponding to morphological analyses $m$ are denoted by $\mathbf{m}$. The embedding vectors corresponding to category-value pairs $f$ are denoted by $\mathbf{f}$.

\subsection{Analysis Embeddings}

\begin{figure}[t]
\centering
\includegraphics[width=\textwidth]{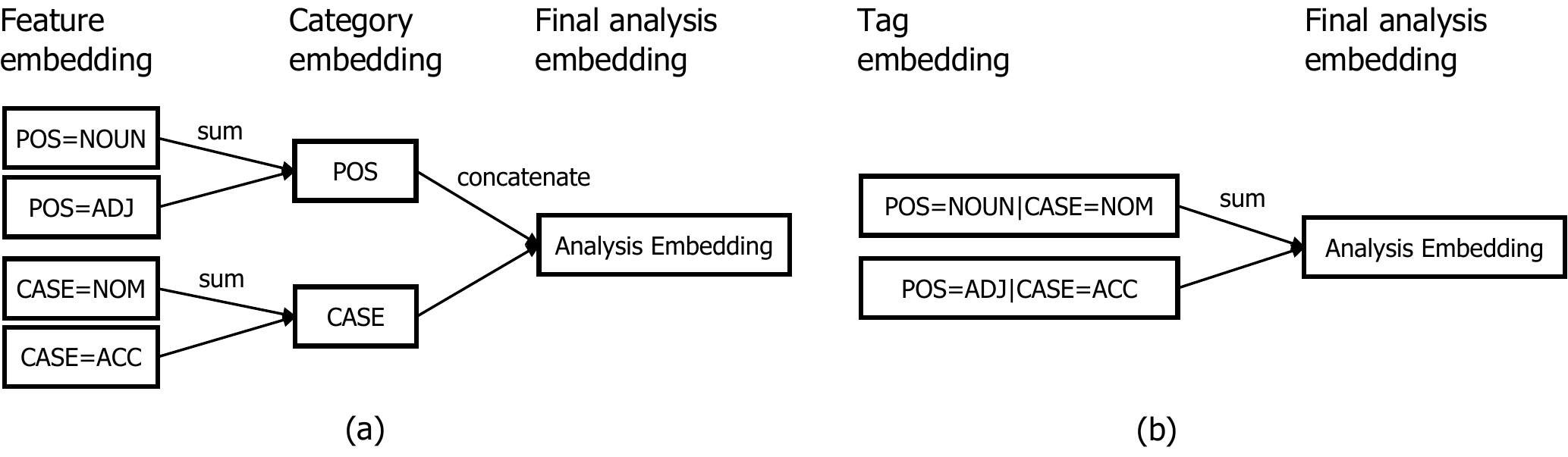}
\caption{Constructing an analysis embedding for a word with \textsc{MA} analyses \textsc{[POS=Noun,CASE=Gen]} and \textsc{[POS=Adj,CASE=Acc]} using (a) category-based and (b) tag-based approach.}
\label{fig:ma_embed}
\end{figure}

The Analysis Embeddings approach is inspired by Inoue et al.~\cite{inoue2017}. Here, we directly augment the input word representations with the dense representations of the morphological analyses generated by the \textsc{MA}.

We examine two methods to compute the analysis embedding for a word -- a tag-based method and a category-based method. 
In the category-based method, which was introduced by Inoue et al.~\cite{inoue2017}, we decompose morphological analyses into a list of individual key-value features $f$ (e.g. \textsc{POS=Noun})  and learn vector representations $\mathbf{f}$ for each such feature.
For each feature key (e.g. \textsc{POS}), we first compute the category embedding as a sum of the embeddings of the corresponding values contained in the \textsc{MA} analyses (see Figure~\ref{fig:ma_embed} (a)).
The final analysis embedding for a word is obtained by concatenating individual category embeddings in a pre-fixed order. The analysis embedding is then concatenated to the word and character-based embedding to obtain an input representation fed into the encoder.



The category-based method assumes that the morphological tags must be given as a list of category-value pairs, or at least is should be possible to convert the tags into such format relatively easily. While this is the case for the Estonian resources we are using in this work, this assumption might not necessarily hold for morphological analysers or lexicons in other languages. Thus, for the sake generality, we also explore a simpler method that does not make any assumptions about the format of the morphological tags.
In the tag-based method, we treat each \textsc{MA} analysis as a single combined tag $m$ and simply encode them using dense vector representations $\mathbf{m}$.
The analysis embedding for a word is then computed as a sum of individual \textsc{MA} tag embeddings (see Figure~\ref{fig:ma_embed} (b)).


\subsection{Attention Embeddings}

\begin{figure}[t]
\centering
\includegraphics[width=\textwidth]{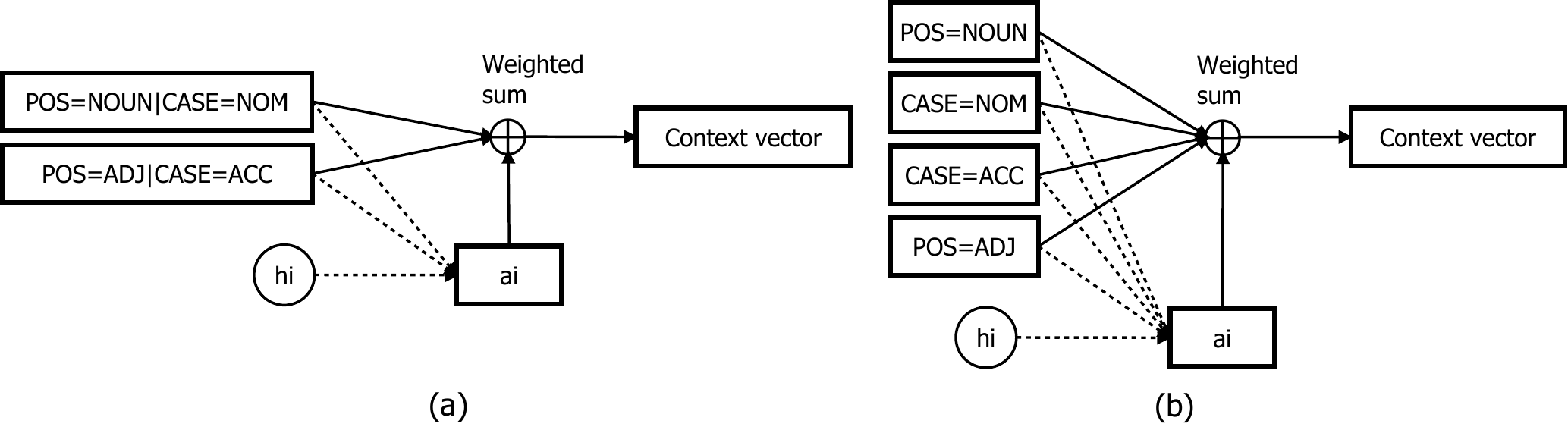}
\caption{Constructing an attentional context vector for a word with analyses \textsc{[POS=Noun,Case=Gen]} and \textsc{[POS=Adj,Case=Acc]} using (a) tag-based and (b) category-based approach.}
\label{fig:ma_atn}
\end{figure}

In the Analysis Embeddings approach, the augmented word representations are obtained by combining all MA analyses, both correct and wrong, with equal weights.
Alternatively, an attention-based method \cite{Luong2015,Bahdanau2014} would enable to weigh MA analyses differently, such that the most relevant analyses will have greater contribution to the final tagging decision. 
To the best of our knowledge, the attempt to utilise attention to guide the neural morphological tagging/disambiguation is novel.

Similarly to Analysis Embeddings method, attentional context vectors  can be computed in two distinct ways -- based on the full \textsc{MA} analysis tag and based on individual categories as illustrated in Figure~\ref{fig:ma_atn}. 
In the tag-based approach, the attention is computed over analysis embedding vectors $\mathbf{m}$. The unnormalised attention scores between the $i$th word and the $j$th tag are computed according to the Luong's global attention \cite{Luong2015}. The scores are normalised via softmax and the context vector $\mathbf{c}_i$ is computed as a weighted average over the tag embedding vectors. 
The context vector $\mathbf{c}_i$ is concatenated with the latent feature vector $\mathbf{h}_i$, passed through a dense layer with a non-linear activation to obtain an attention-informed feature vector $\mathbf{\bar{h}}_i$. The attention-informed feature vector $\mathbf{\bar{h}}_i$ is passed through a linear layer and then through a softmax to compute the predictive probabilities of the tags.
The category-based attention is applied similarly. However, instead of the analysis embeddings $\textbf{m}$, the attention is applied over the set of distinct  category-value embeddings $\textbf{f}$ obtained from all morphological analyses generated to the word by the MA.

The application of the attention to the MA outputs differs for the \textsc{Mc} and \textsc{Seq} models.
In the case of the \textsc{Mc} model, we compute attention alignments with respect to the encoder outputs $\mathbf{h}_i$. 
In the case of the \textsc{Seq} model, the attention is applied at each step when generating a label to a word. 
Specifically, we compute the alignments with respect to MA analyses at each decoding step when predicting each morphological category-value pair. 

%% file: experiments.tex
\section{Experimental Setup}
\label{sec:experiments}

\paragraph{Datasets}
We evaluate our models on three datasets: Morphologically Disambiguated corpus (\textsc{MD})\footnote{http://www.cl.ut.ee/korpused/morfkorpus}, Estonian Dependency Treebank (\textsc{EDT})\footnote{https://github.com/EstSyntax/EDT} and Estonian Universal Dependencies version 2.2 (\textsc
{UD})\footnote{https://github.com/UniversalDependencies/UD_Estonian-EDT/}.
We transform morphological annotations in the \textsc{MD} and the \textsc{EDT} datasets into a \textsc{UD}-like key-value format by organising individual morphological attributes into groups of morphological categories.
For model training and evaluation purposes, we split the \textsc{MD} and \textsc{EDT} datasets sentence-wise into training, dev and test sets with a ratio of 80\%, 10\% and 10\%, while the \textsc{UD} dataset comes with the pre-defined split.
Table~\ref{tbl:corpus_stats} provides summary statistics for all datasets.

\begin{table}[t]
\centering
\footnotesize
\caption{Summary statistics for our datasets.}
\setlength\tabcolsep{5pt}
\begin{tabular}{lrrr | rrr | rrr}
\toprule
                & \multicolumn{3}{c|}{\textsc{MD}} & \multicolumn{3}{c|}{\textsc{EDT}} & \multicolumn{3}{c}{\textsc{UD}} \\
 & train & test & dev & train & test & dev & train & test & dev \\
\midrule
Tokens          & 499 273 & 62 655 & 62 649 & 345 463 &  43 058 & 43 145 & 287 859 & 41 273 & 37 219 \\
Types           &  77 275 & 19 354 & 18 982 &  68 714 &  15 949 & 16 000 &  58 942 & 12 613 & 12 972 \\
Tags            &     385 &    299 &    315 &     719 &     471 &    474 &     918 &    559 &    580 \\
\bottomrule
\end{tabular}
\label{tbl:corpus_stats}
\end{table}

\paragraph{Vabamorf}
We use \textsc{Vabamorf} morphological analyser to obtain candidate analyses for soft neural disambiguation.
We run \textsc{Vabamorf} with a guesser, proper name resolver and disambiguator.
Since some of our models utilise \textsc{Vabamorf}'s outputs on category-level, we convert them to the key-value format as explained above.

\paragraph{Word Embeddings}
We use \textit{fastText} word embeddings\footnote{https://github.com/facebookresearch/fastText/blob/master/pretrained-vectors.md}~\cite{bojanowski2017}.
The dataset contains 329~987 300-dimensional word vectors trained on Estonian Wikipedia.
Although these embeddings are uncased, our model still captures case information by means of character-level embeddings.
Words in a training set with no pre-trained embeddings are initialised with random embeddings. 
At test time, words with no pre-trained embedding are assigned a special UNK-embedding.
We train the UNK-embedding by randomly substituting singletons in a batch with the UNK-embedding with a probability of 0.5.  

\paragraph{Training and Parametrisation}
We train all neural models using stochastic gradient descent for up to 400 epochs and stop early if there has been no improvement on development set within 50 epochs.
The batch size is set to 5 for \textsc{Seq} models and 20 for \textsc{Mc} models.
The initial learning rate is set to 1 for both models.
For \textsc{Mc} model, we decay the learning rate by a factor of 0.98 after every 2500 batch updates.
We represent input words with 300-dimensional word-level embeddings and 150-dimensional character-level embeddings.
The encoder employs a single-layered biLSTM with a 400-dimensional hidden state.
We apply a dropout of 0.5 on both input layer and encoder outputs.
In the Analysis Embeddings approach, we use 50-dimensional tag-embeddings and 10-dimensional category-embeddings, while in the Attention Embeddings approach, both category- and tag-embeddings are 50-dimensional.
We initialise biases with zeros and parameter matrices using Xavier uniform initialiser.

%% file: results.tex
\section{Results}
\label{sec:results}

\begin{table}[t]
\caption{\small{Morphological tagging accuracies for our neural models (a) and baselines (b). The best results are marked in bold.}}
\label{tbl:results_tag}
    \begin{subtable}[t]{.5\linewidth}
      \caption{Neural models}
      \centering
        \begin{tabular}{lrrr}
\toprule
\textsc{Model} & MD & EDT & UD \\
\midrule
\textsc{Mc}	            & 97.00	& 95.60	& 94.43 \\
\textsc{Mc+Emb-tag}	    & 98.02	& 96.01	& 95.01 \\
\textsc{Mc+Emb-cat}	    & \textbf{98.03}	& \textbf{96.06}	& 95.02 \\
\textsc{Mc+Atn-tag}     & 97.80	& 95.61 & 94.59 \\
\textsc{Mc+Atn-cat}     & 97.82	& 95.61 & 94.72 \\
\midrule
\textsc{Seq}	        & 96.85	& 95.37	& 94.32 \\
\textsc{Seq+Emb-tag}	& 98.00	& 95.92	& \bf 95.19 \\
\textsc{Seq+Emb-cat}	& 97.96	& 95.93	& \textbf{95.19} \\
\textsc{Seq+Atn-tag}	& 97.94	& 95.75	& 94.78 \\
\textsc{Seq+Atn-cat}	& 97.95	& 95.67	& 94.71 \\
\bottomrule
        \end{tabular}
    \end{subtable}%
    \begin{subtable}[t]{.5\linewidth}
      \centering
        \caption{Baselines}
        \begin{tabular}{lrrr}
\toprule
\textsc{Model} & MD & EDT & UD \\
\midrule
\textsc{VabaMorf}   & 90.22 & 79.17 & - \\
\textsc{MarMoT}     & 95.23 & 93.01 & 92.03 \\
\textsc{Mc} \cite{heigold2017} & - & - & 94.25 \\
\textsc{\textsc{Mc } \cite{tkachenko2018}}& - & - & 93.28\\
\textsc{\textsc{Seq} \cite{tkachenko2018}} & - & - & 93.30\\
\bottomrule
        \end{tabular}
    \end{subtable} 
\end{table}

\begin{table}[t]
\caption{Performance of \textsc{Mc} and \textsc{MC+EMB-CAT} models on individual features reported as macro-averaged F1-scores. Table rows are sorted by feature frequency in the \textsc{MD} dataset.}
\centering
\footnotesize
\setlength\tabcolsep{4pt}
\begin{tabular}{lccr | ccr| ccr}
\toprule
    & \multicolumn{3}{c|}{\textsc{MD}} & \multicolumn{3}{c|}{\textsc{EDT}} & \multicolumn{3}{c}{\textsc{UD}}\\
Feature & \textsc{Mc} &\textsc{+Emb-Cat} & diff & \textsc{Mc} & \textsc{+Emb-Cat} & diff & \textsc{Mc} & \textsc{+Emb-Cat} & diff \\
\midrule
POS & 93.34 & 95.12 & +1.78 & 90.54 & 91.17 & +0.63 & 87.83 & 88.18 & +0.35 \\
Number & 98.73 & 99.16 & +0.43 & 98.42 & 98.86 & +0.44 & 98.06 & 98.52 & +0.46 \\
Case & 96.05 & 98.31 & +2.26 & 95.25 & 95.52 & +0.27 & 94.46 & 95.26 & +0.80 \\
NounType & 97.38 & 98.40 & +1.02 & 97.15 & 97.62 & +0.47 & - & - & - \\
Punct & 100.00 & 100.00 & 0.00 & 96.49 & 96.60 & +0.11 & - & - & - \\
VerbType & 96.96 & 97.39 & +0.43 & 97.26 & 97.37 & +0.11 & - & - & - \\
Voice & 97.62 & 98.02 & +0.40 & 96.33 & 96.80 & +0.47 & 98.68 & 98.88 & +0.20 \\
Tense & 97.44 & 97.56 & +0.12 & 96.82 & 97.10 & +0.28 & 93.62 & 93.70 & +0.08 \\
Polarity & 99.03 & 99.29 & +0.26 & 99.15 & 99.32 & +0.17 & 99.30 & 99.22 & -0.08 \\
Mood & 95.93 & 99.07 & +3.14 & 93.64 & 95.80 & +2.16 & 87.82 & 89.99 & +2.17 \\
Person & 97.65 & 98.56 & +0.91 & 97.22 & 97.80 & +0.58 & 96.04 & 96.06 & +0.02 \\
Degree & 96.98 & 98.41 & +1.43 & 95.90 & 96.02 & +0.12 & 95.39 & 95.93 & +0.54 \\
VerbForm & 97.41 & 98.07 & +0.66 & 96.82 & 97.00 & +0.18 & 98.53 & 98.59 & +0.06 \\
NumForm & 97.33 & 97.91 & +0.58 & 89.68 & 91.94 & +2.26 & 84.20 & 87.08 & +2.88 \\
NumType & 97.87 & 98.53 & +0.66 & 95.59 & 96.22 & +0.63 & 91.25 & 91.69 & +0.44 \\
Abbr & 95.78 & 95.34 & -0.44 & 97.58 & 98.22 & +0.64 & 89.94 & 89.58 & -0.36 \\
PronType & - & - & - & 95.35 & 97.28 & +1.93 & 89.97 & 89.25 & -0.72 \\
AdpType & - & - & - & - & - & - & 89.28 & 88.94 & -0.34 \\
Connegative & - & - & - & - & - & - & 99.10 & 98.76 & -0.34 \\
Poss & - & - & - & - & - & - & 95.87 & 95.44 & -0.43 \\
Reflex & - & - & - & - & - & - & 93.68 & 92.55 & -1.13 \\
Hyph & - & - & - & - & - & - & 99.03 & 98.00 & -1.03 \\
\midrule
Average & 97.22 & 98.07 &  & 95.83 & 96.51 &  & 93.79 & 93.98 &  \\
\bottomrule
\end{tabular}
\label{tbl:results_feature}
\end{table}

%

Table \ref{tbl:results_tag} (a) reports full-tag accuracies on the \textsc{MD}, \textsc{EDT} and \textsc{UD} test sets.
In addition to our neural models, we also report results for \textsc{Vabamorf} MA\footnote{Ambiguous words are resolved by always picking the first analysis.} and for \textsc{MarMoT}, a CRF-based morphological tagger
\cite{mueller2013} (b).
We also include recent results by Heigold et al.~\cite{heigold2017} and Tkachenko and Sirts~\cite{tkachenko2018} which were, however, obtained on older and smaller versions of the \textsc{UD} dataset, and thus are not directly comparable to ours.
Results indicate that both our basic neural models -- \textsc{Mc} and \textsc{Seq} --  outperform \textsc{MarMoT} and \textsc{Vabamorf} baselines\footnote{The UD annotation is very different from the annotation used by \textsc{Vabamorf} and thus a reliable evaluation of \textsc{VabaMorf} on \textsc{UD} corpus could not be made.} by a large margin while the \textsc{Mc} model demonstrates consistent improvement over the \textsc{Seq} model on all datasets.
The latter finding contradicts Tkachenko and Sirts~\cite{tkachenko2018} who argue in favour of the \textsc{Seq} model.
This can be explained by the fact that in~\cite{tkachenko2018}, the authors train models using a significantly smaller dataset
which contains a larger number of infrequent and OOV tags to which the \textsc{Mc} model is especially susceptible.

Incorporating \textsc{MA} outputs further boosts performance for both models from 0.46\% (\textsc{Mc-Emb-Cat} on EDT) to 1.15\% (\textsc{Seq-Emb-Tag} on MD).
We hypothesised that Attention Embeddings would perform better than Analysis Embeddings because by using Attention Embeddings, the model can learn different weights to input analyses depending on context. Interestingly, this hypothesis did not hold and the Analysis Embeddings method performs consistently better in all cases.
The difference between the category- and tag-based approach is, however, negligible, suggesting that it is not necessary to convert the MA analyses into category-value format.


Table~\ref{tbl:results_feature} compares the performance of the \textsc{Mc} and \textsc{Mc+Emb-Cat} models on individual features.
On MD and EDT corpora, the prediction accuracy improves for all categories. The only exception is the abbreviation category, for which the F1-score drops a little. 
For MD corpus, adding MA outputs helps most to improve the prediction performance of \textsc{Mood}, \textsc{Case} and \textsc{Degree} categories. The accuracy of the \textsc{POS} as well as the \textsc{NounType} (which essentially also encodes POS information) also improves considerably. For EDT corpus, the categories \textsc{NumForm}, \textsc{Mood} and \textsc{PronType} improve the most. The UD corpus gains the least from the addition of MA outputs. In fact, the macro-averaged F1-scores of 8 categories actually decrease. One reason for these results might be that the UD annotations differ the most from \textsc{Vabamorf} annotations. More detailed analysis of these results is left for future work. However, we note that even supplying MAs with very different annotations, the neural tagger is able to improve the overall tagging performance on the UD corpus as well.


%% file: conclusion.tex
\section{Conclusion}
\label{sec:conclusion}

In this work, we focused on Estonian morphological tagging using neural sequence tagging framework.
We experimented with two neural architectures which differ in the way they model morphological tags.
Our results show that both neural models demonstrate high performance, consistently outperforming both rule-based \textsc{Vabamorf} and a CRF-based baseline on three datasets.
We further explored multiple ways to utilise word analyses produced by a morphological analyser as additional inputs to our neural models.
We found that in all cases, the augmented models achieved a consistently superior performance on our datasets, including \textsc{UD} which uses a considerably different tagset than the \textsc{Vabamorf} MA.
The highest improvement for both models was achieved by incorporating morphological analyser outputs in the form of category-level embeddings as inputs to the tagger encoder.

%% file: acknowledgments.tex
\section*{Acknowledgments}
This work was supported by the Estonian Research Council (grants no. 2056 and 1226).